\documentclass[11pt]{article}
\usepackage{eamt26}
\usepackage{times}
\usepackage{url}
\usepackage{latexsym}
\usepackage[small,bf]{caption}
\usepackage{booktabs}
\usepackage{multirow}
\usepackage[utf8]{inputenc}
\usepackage{longtable} % <-- toevoegen
\usepackage{booktabs}
\usepackage{hyperref}
\usepackage{float}
\usepackage[authoryear]{natbib} 

\title{Metaphors in Literary Post-Editing: Opening Pandora's Box?\\
}
\author{
Aletta G. Dorst, 
Mayra O. Nas and 
Katinka Zeven\\ Leiden University Centre for Linguistics\\
{\tt \{a.g.dorst, m.o.nas, k.zeven\}@hum.leidenuniv.nl}
}

\date{}
\begin{document}
\maketitle

\begin{abstract}
This paper investigates how post-editors of literary texts react and respond to the way metaphors have been translated by Neural Machine Translation (NMT) and Large Language Models (LLMs). The results show that one in three metaphors in the output were changed by the post-editors, demonstrating that the translation of figurative language is indeed problematic in literary MT (LitMT). The responses indicate that the post-editors were aware of overly literal translations, though mostly for multiword expressions. Moreover, at times they found it difficult to determine whether solutions were acceptable. They rated the overall quality of the MT output as quite poor and stated that the post-editing was more work and more effort than it would have been translating from scratch. This supports previous studies arguing that post-editing constrains translators in their creativity and diminishes their sense of text ownership.
\end{abstract}

\section{Introduction}
LitMT is no longer a fantasy, a perhaps equally fascinating and frightening scenario of science fiction. In 2023, Reedz was heralded as “the world's first fully AI powered publisher” \citep{reedz2023} and Nuanxed has been making a name for itself since 2021 by turning AI post-editing into its standard workflow to “give books the audience they deserve” by making “quality book translations […] easy, fast and affordable”\footnote{https://www.nuanxed.com/}. The same positive, almost idealistic promise of providing authors with fair and affordable access to new audiences and helping them fulfill their dreams of fame and fortune was found in Amazon’s launch of Kindle Translate, “an AI-powered translation service for authors to reach global readers” creating opportunities for authors “to reach new audiences and earn more”\footnote{https://www.aboutamazon.com/news/books-and-authors/amazon-kindle-translate-books-authors}.

Despite this alluring promise, literary authors and translators alike remain skeptical about the suitability of machine translation (MT) and post-editing (PE) for literary texts \citep{moorkens2018, way2023}. While research shows that the quality of MT output has improved considerably since the introduction of neural techniques \citep{bentivogli2016,castilho2017}, it has mostly been a matter of marketing hype to state that neural MT (NMT) has “bridged the gap” between human and machine translation \citep{wu2016} or achieved “human parity” \citep{hassan2018-xu}, as shown by, amongst others \citet{laubli2018-tq} and \citet{toralway2018}. Even state-of-the-art NMT and LLM translation produces output that contains errors and cannot be published without human revision. As pointed out by \citet[p.~89]{way2023}, “claims that human translators are doomed as their jobs will be taken over completely by machines” and “[o]verhyping the capability of the technology does nobody any good”. At the same time, the authors argue that the time has come to challenge the assumption that Computer-Assisted Translation (CAT), including MT, is “too crude and wayward to be of service” (\textit{ibid}) in the translation of literary texts.

The current study advances the debate surrounding literary MT and post-editing by focusing specifically on how post-editors react and respond to the way metaphors have been translated by NMT and LLMs in three short excerpts from a spy novel. We focus on metaphors because they are a well-known problem in translation and previous research indicates that MT has a tendency to translate metaphors too literally \citep{zajdel2022, dorst2023, karakanta2025}, “leading to nonsensical, word-for-word translations” \citep[p.~134]{zajdel2022}. Metaphors are thus a persistent problem in MT that requires human post-editing. Our investigation aims to shed more light on which metaphors are perceived as problematic and how post-editors handled the way machines translate them. 

\section{Related Work}

\subsection{LitMT}
Although literary translation was for a long time considered “the last bastion of human translation" and a challenge to “the perceived wisdom [...] that MT is of no use for the translation of literature" \citep[p.~123]{toral2015b}, the examples in the Introduction show that the publishing industry is ready to embark full speed ahead. And while researchers warn against overstating MT's capabilities (e.g. \citealp{laubli2018-tq,toral2018a,way2023}), both in industry and academia, a growing number of studies has explored the potential of statistical and neural systems in handling different literary genres across different language pairs (e.g.\ \citealp{besacier2014,genzel2010,greene2010,jones2013,ploeger2024,thai2022,toral2023,toral2015a,toral2015b,toralway2018,voigt2012}).

Studies assessing the potential of MT for particular genres and domains typically measure the quality of the MT output using automatic metrics, such as BLEU \citep{papineni2002}, METEOR \citep{lavie2007} and COMET \citep{rei2020}. As pointed out by \citet{docarmo2022}, such measurements often start from a misleading conceptualization of the notion of “quality”; moreover, the ensuing misplaced confidence that these metrics enable them to reliably measure quality prevents scholars from reflecting critically on their usefulness. In a similar vein, \citet{vanegdom2023} argue that “metrics and algorithms cover only parts of the notion of “quality”, and that a more fine-grained approach is needed if potential literary quality of machine translation is to be captured” (p. 129).

At present, most studies assessing the quality of literary MT output employ a combination of automatic metrics and human evaluation; in those human evaluations, usually carried out by native speakers, readers have been reported to rate a considerably high number of MT sentences as acceptable, error-free or equivalent to human translation: 60\% for Spanish to Catalan \citep{toralway2018}; 34\% for English to Catalan \citep{toral2018a}; \textasciitilde{}20\% for English into Russian and German \citep{matusov2019}; and 44\% for English into Dutch \citep{fonteyne2020}. However, as \citet{vanegdom2023} point out: “evaluative judgments are often passed by individuals who lack the required expertise to judge the quality […] Evaluation is often performed by people that were simply available and willing to help out" (p. 131). Rather tellingly, a recent multilingual study involving 20 language pairs reported that professional translators preferred human translations to MT 85\% of the time \citep{thai2022}.

While recent research convincingly demonstrates that the quality of LitMT output can be significantly improved by techniques including domain adaptation \citep{toral2023}, author-tailored adaptation \citep{kuzman2019,oliver2023} and restoration of lexical richness \citep{ploeger2024}, there is still a sizeable gap between the output and publishable translations, especially in terms of adequacy, style, tone, cohesion and the translation of figurative language \citep{tezcan2019, matusov2019, hansen2022}. Of the different forms of figurative language, linguistic metaphors - especially idioms and multi-word expressions - continue to be a problematic characteristic of literary texts and notoriously hard to translate for machines \citep{dorst2023, karakanta2025, zajdel2022}. Even state-of-the-art NMT and LLMs have a tendency to translate metaphors literally (i.e. word-by-word), resulting in unidiomatic, illogical, even non-sensical translations \citep{dorst2023, karakanta2025, zajdel2022, tezcan2019}. In the current study, we therefore focus on whether literary post-editors notice such overly literal metaphor translations and how they react and respond to them. 

\subsection {Post-editing literary texts} 

Surveys on literary translators’ perceptions and use of technology – ranging from simple online dictionaries and word processing to CAT tools and MT – such as those conducted by \citet{ruffo2018, ruffo2022, ruffo2024} and \citet{daems2021,daems2022a,daems2022b}, consistently indicate that while most literary translators embrace technologies such as the Internet, glossaries and terminology tools, they are often reluctant to use CAT tools and MT for multiple reasons: they have often not received any training in how to use such tools, they are unaware of the latest developments, they consider such tools unsuitable for creative texts, and they are opposed to “tools that threaten to steal the essence of their translation activity, ignoring the peculiarly human aspects of it” \citep[p. 130]{ruffo2018}. 

Interestingly, the study by \citet{moorkens2018} on literary post-editing versus translation 'from scratch' found that while all participants “were faster when post-editing NMT, […] they all still stated a preference for translation from scratch, as they felt less constrained and could be more creative" (p. 256). The participants in the study “complained that the MT systems ‘conditioned' them to produce a literal translation" (\textit{ibid}). 

This sense of being constrained by the MT output and being lured into accepting MT output was also found in a series of experiments conducted by Guerberof-Arenas and colleagues \citep{guerberof2020,guerberof2022,guerberof2024}, focusing specifically on creativity, as well as the real-life case studies conducted by \citet{kenny2020}, \citet{winters2023} and \citet{kolb2023_brazilian, kolb2023_surprised}. These findings are not only relevant from a stylistic perspective, but also raise legal and ethical questions. \citet{taivalkoski2023}, for example, write about the ongoing debate regarding copyright, authorship and intellectual property in translation. They note how post-editing poses a serious challenge to traditional copyright laws as authorship will depend on how much of the raw MT output is kept and how much “originality" is added through the post-editor's unique changes.

In the study by \citet{guerberof2020} human translation scored higher on creativity than MT and PE and ranked higher in narrative engagement and translation reception. Guerberof-Arenas and Toral therefore argue that “professional translators, by providing solutions that are both novel and acceptable, add the creativity factor that MT is lacking at present" (p. 277). Moreover, \citet{guerberof2022} found not only that their literary-adapted neural MT system did not have the “necessary capabilities for a creative translation" but also that “using MT to post-edit raw output constrains the creativity of translators, resulting in a poorer translation often not fit for publication, according to experts" (p. 184). In the same study, the reviewers were unanimous in their verdict that the human translations were good, the MTs were bad, and the post-edited versions were neither good nor bad. 

In the current study we investigate how literary post-editors react and respond to the way metaphors have been translated by MT systems, both NMT and LLM, while post-editing short excerpts from a novel. We are particularly interested in how often metaphor translations were changed by the post-editors and how aware they were of overly literal translations that require more creative solutions.

\section{Methodology}

\subsection{Machine Translations} Three systems were used to generate machine translations for the literary post-editing task: two commercial NMT engines – Google Translate and DeepL – and one general-purpose LLM, ChatGPT (GPT4; accessed via the ChatGPT user interface). All systems were accessed via their online web interfaces \footnote{https://translate.google.com/; https://www.deepl.com/nl/translator; https://chatgpt.com/}, reflecting the way most literary translators use MT, if at all. Most professional literary translators do not build or customize their own MT systems, and when they do post-editing assignments, they often receive the MTs as editable Word files \citep{ruffo2024, macken2025}. ChatGPT received a bare prompt (“Translate the following English text into Dutch for the Netherlands”). No in-domain training, customization or finetuning took place; all translations were generated “out of the box”. These three systems were chosen because they are widely used, freely accessible online, and representative of the engines most translators are familiar with. Three different versions of the PE task were created so each engine was used for one fragment and the post-editors saw one output from each of the three engines.

\subsection{Source Text: The Lucy Ghosts (Eddy Shah)}
The novel used as the source text (ST) in this study was selected from the Fiction subcorpus of the annotated VU Amsterdam Metaphor Corpus (VUAMC) \citep{steen2010corpus}. The annotated VUAMC is available for downloading for non-commercial use via the Oxford Text Archives website \footnote{\url{https://ota.bodleian.ox.ac.uk/repository/xmlui/handle/20.500.12024/2541?show=full}}. All texts in the VUAMC were sampled from the BNC Baby corpus \footnote{https://www.natcorp.ox.ac.uk/corpus/babyinfo.html} with permission under NWO Vici grant 277-30-001 \footnote{https://www.nwo.nl/projecten/277-30-001-0}. All sampled excerpts were manually annotated for linguistic metaphor by a team of experts using the Metaphor Identification Procedure VU or MIPVU \citep{steen2010mipvu}. The VUAMC has been widely used as the gold standard in NLP studies on metaphor processing. Three short excerpts were selected (798 words in total), offering a range of different types of metaphor (i.e. single vs multiword metaphors; different word classes; different degrees of creativity). Since the post-editors were revising three short excerpts instead of the whole text, a synopsis and some background information were provided for contextualization along with the ST excerpts.

\subsection{Post-editors}
In total, six post-editors (PEs) were invited to participate in the task and subsequent interview. All six had recently graduated from the Master's in Translation at Leiden University in the Netherlands and completed the specialization in Literary Translation as well as a course in Translation Technology. This ensured a relatively homogeneous background and a reasonable familiarity with translation technology, including MT. We invited graduates who we knew received high grades for their literary translations and who are currently working as professional translators, both in literary and non-literary translation, with no more than 5 years of professional experience. The participants were told they would complete a literary post-editing task, and would be invited to share their thoughts on the suitability of MT for literary texts, but they were not informed of the authors' particular interest in metaphors. The participants signed an Informed Consent Form before the task and were debriefed during the interview. They received financial compensation for their participation.

\subsection{Data Collection - PE task and Interview}
Each of the PEs came to Leiden to complete the task and interview individually. They were encouraged to bring their own laptop, which three of the six participants did. The other three worked on a laptop provided by the authors. The PEs received two Word documents: one with the three ST fragments (F1, F2, F3) and some information on the novel, the plot and the setting of the excerpts; and one with three MT outputs. There were three versions: PE1 and PE2 received version 1, with F1=DeepL, F2=ChatGPT, F3=GT; PE3 and PE4 received version 2, with F1=GT, F2=DeepL, F3=ChatGPT; and PE5 and PE6 received version 3, with F1=ChatGPT, F2=GT, F3=DeepL). The document also included instructions for the PE task: participants were instructed to work with Tracked Changes and Comments and use any resources they required. A subset of the data and materials associated with this study is publicly available in a GitHub repository \footnote{https://github.com/dorstag/}.

During the post-task interview, two of the three authors were present. The interviews were semi-structured and started from a predefined list of questions. Our interest in metaphors was brought up towards the end of the interview, so we could see whether problems with specific metaphors or figurative language more generally were raised by the post-editors themselves. The interviews were recorded using a smartphone and subsequently transcribed by one of the authors. All interviews were conducted in Dutch; quotations from the PEs were translated into English by the authors for the sake of readability.

Transcribed interviews were uploaded into ATLAS.ti to allow for a more systematic and quantitative analysis of the post-editors' attitudes toward literary PE and the suitability of MT for literary translation. The coding process followed a bottom-up, inductive thematic analysis in line with the approach outlined by \citet{braun2006}. This iterative process allowed the analysis to evolve from general themes towards more concrete conceptual groupings \citep[p. 12-13]{skjottlinneberg2019}. To prevent the unconscious falsification of data by coding it according to theoretically desirable patterns \citep[p. 127]{schuyt2019}, we annotated and reviewed the dataset both separately as well as together and did so multiple times. 

Reflections often included statements that fit multiple categories, and each relevant statement was coded separately. These characteristics should be understood as themes, meaning the post-editors did not have to use the exact wording of the category labels (i.e., if a post-editor said “\textit{aftrap} refers to football, we don't say that in Dutch", a code was added that the PE noticed the metaphor and that they thought it was a bad translation).

\section{Results} This section presents the results of the post-editing task and follow-up interviews with the PEs (hereafter referred to as PE1 to PE6). Our aim was to determine how the PEs reacted to MT metaphors - that is, whether they retained or changed the output - and how they responded to them, focusing on whether their reported perceptions correspond to their actual post-editing behavior.

\begin{table}[H]
\centering
\resizebox{\columnwidth}{!}{
\begin{tabular}{lcccc}
\hline
Engine & F1 & F2 & F3 & Total (mean) \\
\hline
Google Translate & 181 & 184 & 158 & 523 (174) \\
DeepL & 170 & 214 & 169 & 553 (184) \\
ChatGPT & 130 & 205 & 190 & 525 (175) \\
\hline
Total (mean) & 481 (160) & 603 (201) & 517 (172) & 1601 \\
\hline
\end{tabular}
}
\caption{Number of edits per MT system and fragment.}
\label{tab:edits}
\end{table}

To determine whether the MT system influenced editing behavior, the number of edits was analysed per system (Table~\ref{tab:edits}).

 The differences between the three systems are relatively small: DeepL required the highest total number of edits (553), followed by ChatGPT (525) and Google Translate (523). This suggests that there were no clear quality differences between the engines used in the task. How many edits were made seemed to depend more on the ST characteristics (fragment variation) and post-editing style (PE variation) than on which engine was used to generate the output. The strongest PE variation was observed between PE1, who had 389 edits, and PE2, who made 144 edits. The other four fell between these two extremes (see Table~\ref{tab:combined_edits}).

\begin{table*}[!ht]
\centering
\centering
\small
\setlength{\tabcolsep}{4pt}
\begin{tabular}{lccccccc}
\hline
Category & PE1 & PE2 & PE3 & PE4 & PE5 & PE6 & Total (mean) \\
\hline
Google Translate & 122 & 36 & 99 & 82 & 90 & 94 & 523 (87) \\
DeepL & 120 & 50 & 122 & 92 & 89 & 80 & 553 (92) \\
ChatGPT & 147 & 58 & 99 & 91 & 54 & 76 & 525 (88) \\
\hline
F1 & 120 & 50 & 99 & 82 & 54 & 76 & 481 (80) \\
F2 & 147 & 58 & 122 & 92 & 90 & 94 & 603 (101) \\
F3 & 122 & 36 & 99 & 91 & 89 & 80 & 517 (86) \\
\hline
Total (mean) & 389 (130) & 144 (48) & 320 (107) & 265 (88) & 233 (78) & 250 (83) & 1601 \\
\hline
\end{tabular}

\caption{Overview of edits across systems, fragments, and post-editors, including totals per category.}
\label{tab:combined_edits}
\end{table*}

The interviews indicate that these differences do indeed reflect individual post-editing styles, as well as varying levels of experience with professional post-editing, views on the suitability of MT and PE for literature, and personal quality thresholds. Participants reported varying degrees of experience with both MT and PE. While all participants were familiar with MT, either from their studies or their professional work, only half reported having professional post-editing experience. These varying levels of familiarity with MT and (literary) PE may in part explain the observed variation in PE behavior.

\subsection{Reactions to metaphor in MT: Changes made during the post-editing task}

This section focuses on the post-editors’ reactions to metaphor by determining whether they reacted by retaining or changing the MT output.

Table~\ref{changes_metaphor} gives the total number of changes made to the metaphor translations per fragment for each of the six PEs, regardless of engine.

\begin{table}[h]
\centering

\resizebox{\columnwidth}{!}{
\begin{tabular}{lccccccc}
\hline
Fragment & PE1 & PE2 & PE3 & PE4 & PE5 & PE6 \\
\hline
F1 (20) & 11 (55\%) & 6 (30\%) & 9 (45\%) & 7 (35\%) & 8 (40\%) & 7 (35\%)\\

F2 (29) & 17 (59\%) & 10 (34\%)  & 14 (48\%) & 9 (31\%) & 14 (48\%) & 10 (34\%)\\

F3 (20) & 13 (65\%) & 9 (45\%) & 6 (30\%) & 10 (50\%) & 9 (45\%) & 8 (40\%)\\
\hline
Total (69) & 41 (59\%) & 25 (36\%) & 29 (42\%) & 26 (38\%) & 31 (45\%) & 25 (36\%)\\
\hline
\end{tabular}
}
\caption{Number of changes to machine-translated metaphors per fragment and post-editor.}
\label{changes_metaphor}
\end{table}

For all PEs, more than a third of the machine-translated metaphors were problematic, in all three fragments. This indicates that metaphor translation is not a “solved problem” yet, with at least one out of every three metaphors requiring a revision. Considering that corpus-based metaphor studies by
Steen and colleagues ~\citeyearpar{steen2010mipvu,steen2010corpus, steen2010pragglejaz, steen2010usage} have shown that, on average, one in eight words is metaphorical in authentic discourse (including fiction, journalism and academic texts), this entails that post-editors will need to pay careful attention to metaphor translation when working with machine-translated texts.

While there is some variation between the PEs, all of them changed more than 30\% of the metaphors in the fragments (so at least 21 of the 69 metaphors). PE1 has, again, consistently made considerably more changes than all other PEs, changing more than half of the metaphor translations in each fragment. Comparing these findings for changes to metaphor to the previously established overall patterns, we can now see that F2 was problematic for other reasons besides the presence of (difficult) metaphors. 

While all PEs made considerably more revisions in F2, focusing only on how often the metaphors were changed shows more variation across fragments and post-editors: F3’s metaphor translations were considered most problematic by PE1 (65\% changed), PE2 (45\% changed), PE4 (50\% changed) and PE6 (40\% changed) while PE3 and PE5 made more changes to the metaphors in F2 (48\% each). However, PE4 and PE6 made the fewest changes to F2, and PE3 made the fewest changes to F3. This suggests that not all post-editors find the same metaphor translations problematic. 

Interestingly, the few instances in which all six PEs changed the metaphor translation were cases of multiword metaphors where all three systems had produced overly literal translations, as illustrated in Table~\ref{tab:idiom} (see Appendix B for literal word-by-word translations into English). However, the revisions show such considerable variation with respect to \textit{how} the PEs solved the problem that a detailed analysis of their solutions and whether the revision was required or optional is unfortunately beyond the scope of this paper.  

\begin{table}[h]
    \centering
    \includegraphics[width=1\linewidth]{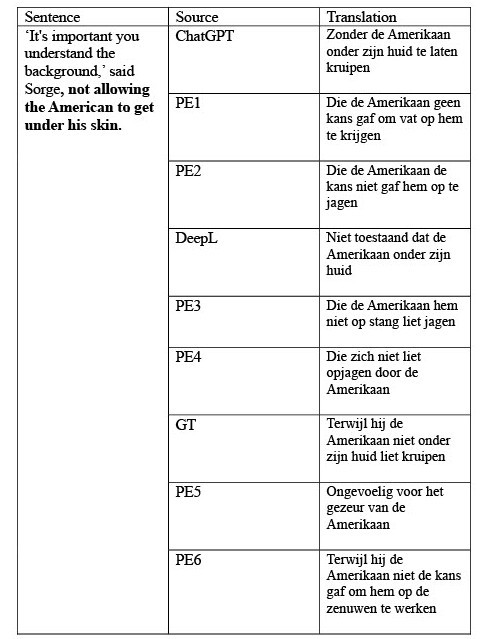}
    \caption{All versions of the idiom “get under his skin”}
    \label{tab:idiom}
\end{table}

\begin{table}[h]
\centering

\resizebox{\columnwidth}{!}{
\begin{tabular}{lccccccc}
\hline
System & PE1 & PE2 & PE3 & PE4 & PE5 & PE6 \\
\hline
GoogleTranslate & 13 (32\%) & 9 (36\%) & 9 (31\%) & 7 (27\%) & 14 (45\%) & 10 (40\%)\\

DeepL & 11 (27\%) & 6 (24\%)  & 14 (48\%) & 9 (35\%) & 9 (29\%) & 8 (32\%)\\

ChatGPT & 17 (41\%) & 10 (40\%) & 6 (21\%) & 10 (38\%) & 8 (26\%) & 7 (28\%)\\
\hline
Total & 41 & 25 & 29 & 26 & 31 & 25\\
\hline
\end{tabular}
}
\caption{ Number of changes to machine-translated metaphors per engine by post-editor.}
\label{tab:pe_fragment}
\end{table}

If we split the data by engine, regardless of fragment, the results per post-editor are as presented in Table~\ref{tab:pe_fragment}. This shows that there is not one engine that produces more problematic metaphor translations overall, as far as the number of revisions made by the post-editors is concerned. PE5 and PE6 made the most changes to the metaphor translations in the Google Translate output, PE3 in the DeepL output, and PE1, PE2 and PE4 in the ChatGPT output. One issue for future investigation is whether the engines did differ systematically in what kind of errors they produced for these metaphor translations. \citet{karakanta2025} showed that NMT made more form errors (fluency) while LLMs made more meaning errors (accuracy). Similarly, the study by \citet{tezcan2019} suggests that metaphors like ‘sport' to refer to a person resulted in errors that were classified both as lexical as well as logical mistakes. 
 
 Further investigation is also needed to tell us whether the changes made by the PEs were required or optional, and whether the MT metaphor translations were in fact “correct” or “incorrect” as classified by an error analysis such as the one carried out by \citet{tezcan2019}. Going through the data to catalogue changes and retentions, we noticed instances where the post-editors have retained metaphor translations that are technically “incorrect” or changed “correct” metaphor translations. For example, PE3 retained DeepL's incorrect translation \textit{‘breken van cyphers}' for ‘breaking cyphers' (lexical error for ‘break' and non-existent word for ‘cyphers'). Conversely, PE1 changed ChatGPT's correct translation \textit{‘aanzienlijk'} (‘considerable'/‘substantial'/‘notable') for ‘substantial' into \textit{‘ongelooflijk succesvol'} (‘incredibly successful'). 

\begin{table}[h]
\centering
\resizebox{\columnwidth}{!}{
\begin{tabular}{lccccccc}
\hline
Met Type & PE1 & PE2 & PE3 & PE4 & PE5 & PE6 \\
\hline
Single (41) & 21 (51\%) & 10 (24\%) & 12 (29\%) & 11 (27\%) & 15 (37\%) & 10 (24\%)\\

Multiword (28) & 20 (71\%) & 15 (54\%)  & 17 (61\%) & 15 (54\%) & 16 (57\%) & 15 (54\%)\\
\hline
Total (69) & 41 (59\%) & 25 (36\%) & 29 (42\%) & 26 (38\%) & 31 (45\%) & 25 (36\%)\\
\hline
\end{tabular}
}
\caption{ Number of changes to machine-translated metaphors per metaphor type by post-editor.}
\label{tab:metaphor_fragment_single_multi}
\end{table}

 Of the 69 metaphors in the three fragments, 41 were single metaphors (e.g. \textit{‘snapped', ‘bypass', ‘sources'}) and 28 were multiword metaphors (e.g. \textit{‘take off’, ‘broke their word’, ‘your hands were not that clean’}). Table~\ref{tab:metaphor_fragment_single_multi} shows that while there is considerable variation between the post-editors in how many of the machine-translated metaphors they changed overall, ranging from a third to over half of all the metaphors, they consistently changed more translations of multiword metaphors than single metaphors. Four of the PEs changed about one in four of the single metaphors, PE5 a third, and only PE1 more than half. For the multiword metaphors, all six PEs changed more than half of the MT translations, and PE1 even changed 20 out of 28, more than 70\%. 
 
 This shows that while multiword metaphors may be less frequent in naturally occurring discourse (and perhaps more strongly related to authorial style, i.e. a preference for idioms or phrasal verbs), they are much more likely to be problematic in MT. Interestingly, all of these multiword metaphors are highly conventional, so their mistranslation by MT cannot be due to obscure or infrequent use. The context shows they are also used in their canonical form. However, taken literally, they evoke rather vivid imagery that is relevant to the story unfolding. One of our next steps is therefore also to determine how readers react and respond to these literal translations and whether they consider them nonsensical or creative.

\subsection{General patterns in the interview data}
As discussed in the previous section, the task revealed differences in post-editing styles. The interview data provides additional insights into these variations. Some participants reflected on their usual PE workflow and reported that they normally adopt a pragmatic approach, preferring to leave acceptable MT output unchanged in order to save time, as they are not compensated sufficiently to deliver perfect quality. PE3 explained this approach as follows:
\begin{quote}
    \textit{Researcher A: Do I understand correctly that, usually, you save time on [post-editing] texts?} 

\textit{PE3: Yes, but then I also think that it's not my best work.} 

\textit{Researcher B: Yeah.}

\textit{PE3: Right? So, as far as I'm concerned, they get a passable translation-- one that's just okay, and that, I think, most people wouldn't consider bad, because I am actually a pretty decent post-editor. But I do believe that it could use an extra round of translating/revising, and that's what I would prefer to do.}
\end{quote}
Similarly, PE6 emphasised that compensation influenced the effort they were willing to invest, focusing on essential corrections only: 

\begin{quote}
    \textit{Researcher A:[...], unless you deliver a lower-quality output.} 
    
\textit{PE6: Yeah, that's definitely what I would do. If I'd get paid less for it [post-editing], I really would... only focus on the vocabulary. And then I'd be like, yeah, figure it out yourself. Yes, ask someone else to do the full editing.} 
\end{quote}

\begin{table*}[!htbp]
\centering
\small
\setlength{\tabcolsep}{4pt} % standaard is ~6pt
\begin{tabular}{lrrrrrrr}
\toprule
Category & PE1 & PE2 & PE3 & PE4 & PE5 & PE6 & Total \\
\midrule
Efficiency: Faster & 0 & 0 & 1 & 0 & 0 & 0 & 1 \\
Efficiency: More work & 4 & 2 & 3 & 4 & 3 & 1 & 17 \\
Efficiency: Slower & 4 & 0 & 0 & 2 & 7 & 2 & 15 \\
Experience: Difficult & 1 & 2 & 1 & 3 & 0 & 0 & 7 \\
Experience: Doubt language skills & 3 & 4 & 2 & 6 & 7 & 0 & 22 \\
Experience: Frustration & 0 & 0 & 0 & 3 & 1 & 0 & 4 \\
Experience: Fun & 1 & 1 & 1 & 0 & 1 & 1 & 5 \\
Experience: Lack context & 2 & 1 & 1 & 1 & 3 & 0 & 8 \\
Experience: Limits creativity & 2 & 1 & 1 & 7 & 2 & 1 & 14 \\
\midrule
Total & 17 & 11 & 10 & 26 & 24 & 5 & 93 \\
\bottomrule
\end{tabular}
\caption{PE Efficiency and Experience}
\label{tab:pe_summary}
\end{table*}

These quotes are clearly aligned with the lack of authorship and text ownership in post-editing discussed by \citet{taivalkoski2023}. Similarly, the quantitative results presented in Table~\ref{tab:pe_summary} align with previous post-editing studies such as those by \citet{moorkens2018} and \citet{ruffo2024}, showing that the participants frequently encountered inefficiencies and challenges during the post-editing task. The ‘Efficiency' categories suggest that, although post-editing was occasionally faster than translating from scratch (e.g. PE3 had a tag in ‘Faster'), a substantial portion of the task required more work or was classified as being significantly slower than translating from scratch, particularly for PE4, PE5 and PE6. 

Similarly, the ‘Experience’ categories reveal that the post-editing task often resulted in difficulties such as self-doubt regarding language proficiency, frustration, and reduced creativity,  among other issues. The issues of doubting their own language skills and feeling constrained in their creativity were often linked in the interviews: the participants described how they could not “unsee" the MT solutions; after having seen the MT solution they could no longer think of alternatives. While they had this nagging suspicion that something was wrong with the translation, they struggled to think how they would have said it themselves and whether it was correct Dutch or not. This confirms previous studies on how PE limits translators' creativity (e.g.\ ~\citealp{guerberof2020,guerberof2022}).

\begin{table}[t]
\centering
\resizebox{\linewidth}{!}{
\begin{tabular}{lrrrrrrr}
\toprule
Metaphor-related code & PE1 & PE2 & PE3 & PE4 & PE5 & PE6 & Total \\
\midrule
Aware after probing        & 0 & 1 & 0 & 0 & 1 & 2 & 4 \\
Mentioned problematic metaphor & 16 & 5 & 1 & 8 & 0 & 2 & 32 \\
Mentioned good metaphor        & 1 & 1 & 0 & 0 & 0 & 1 & 3 \\
Not aware after probing      & 0 & 1 & 1 & 0 & 0 & 1 & 3 \\
\midrule
Total & 17 & 8 & 2 & 8 & 1 & 6 & 42 \\
\bottomrule
\end{tabular}
}
\caption{Metaphor-related quotes}
\label{tab:metaphorquotes}
\end{table}

\begin{table}[t]
\centering
\resizebox{\linewidth}{!}{
\begin{tabular}{lrrr}
\toprule
Category & Total quality mentions & Co-occurring with metaphor\\
\midrule
Bad overall / other     & 18 & 3\\
Grammar                 & 2  & 0\\
Too literal             & 22 & 8\\
Unnatural language      & 19 & 0\\
Vocabulary              & 7  & 0\\
\bottomrule
\end{tabular}
}
\caption{Total mentions of negative quality fragments and their co-occurrence with noticing problematic metaphors}
\label{tab:metaphor_quality_cooccurrence}
\end{table}

\subsection{Responses to metaphor in MT: Reflections on metaphor during the interview}

The interview data provide further insights into whether the participants noticed the metaphors during the PE task, as well as into how the MT translations were perceived. As shown in Table~\ref{tab:metaphorquotes}, 42 out of in total 280 coded segments across nine themes were metaphor-related (see Appendix A for an overview of all coded themes across participants). Of these, the majority (35) consisted of unprompted mentions, indicating that metaphors were frequently noticed spontaneously by the participants. However, there is considerable variation between the PEs: while PE1 mentioned metaphors 17 times, PE5 mentioned them only once, and only after probing by the researchers. This suggests that metaphor awareness differs across individuals. 

As discussed in Section 4.2, PE1 made the highest number of changes to metaphor translations, whereas other PEs, such as PE3 and PE6, changed considerably less. The interview data thus suggest a link between noticing metaphors in the task and mentioning them during the interview: PEs who were more attuned to the metaphorical language were also more likely to revise it.

Several PEs reported issues with unnatural and overly literal language in the texts, as well as errors introduced by the machines (see Table~\ref{tab:metaphor_quality_cooccurrence}). Notably, the researchers purposefully had not yet mentioned metaphors during the part of the interviews when the questions “what did you think of the quality of the machine translations" and “did you notice certain stylistic features" were asked, as illustrated in the quote below.   

\begin{quote}

\textit{Researcher: Did you notice any particular stylistic features?}

\textit{PE3: Let me think- short sentences, repetition, yes, those are things I took into account, of which I thought: this is ugly, but I'll leave it as it is.}

\textit{Researcher: And why did you find it ugly?}

\textit{PE3: Well, because... in some cases it just sounded too strange in Dutch; that you'd think, this is an American expressing things in a certain way and I get that, but it doesn't work like that. So I merged one sentence because of that, and otherwise tried to adapt it... There were simply sentences where I thought: Yes, but this is just not how we would say it.}
\end{quote}

However, references to metaphors frequently co-occurred with comments about overall poor quality and the literalness of the MT output. In eight of the 32 instances in which metaphors were mentioned, the PEs also mentioned that the metaphor was translated too literally. This pattern suggests that by mentioning metaphors in conjunction with excessive literalness or broader issues (as reflected by the dual tagging), the PEs consciously linked the MT output as being too literal and not good enough in terms of metaphor translation. PE4, for example, explicitly connected incorrect metaphor translations to literalness: 
\begin{quote}
    \textit{Researcher A: And the overly literal things? Could you give an [example]?} 
    
    \textit{PE4: Yes, so ‘breaking ciphers' was translated as ‘ciphers breken'.}  
\end{quote}
Here too, the PE connects this problem to the struggle to think of alternatives: 

\begin{quote}
 \textit{PE4: And I also often had the problem that there was something idiomatic in the source text, which, as far as I could tell, had not been correctly translated in... by the machine, but I, because I had already been primed, I found it difficult to think of a solution that would be correct in Dutch.}
\end{quote}

The same problem is raised by PE1: 
\begin{quote}

\textit{Researcher: What we were secretly looking for in these texts was what you did with the literally translated metaphors. So, for example, ‘under your skin'. Did those stand out or not necessarily more than other things?}

\textit{PE4: They definitely stood out, but because I got stuck in that track, it was often hard to find what the correct solution actually was.}
\end{quote}

\section{Conclusion} This study set out to examine how post-editors react and respond to metaphor in literary MT. The findings from the post-editing task and subsequent interviews demonstrate that metaphor translation continues to pose a challenge for both NMT and LLM-based systems. While the general editing patterns demonstrated greater variation between both fragments and PEs, all six participants changed more than a third of the MT metaphors across engines and fragments, indicating that one in three metaphors in MT may be considered problematic. 

One key finding is that this is particularly true for multiword metaphors, which were changed over half of the time by all PEs across fragments and engines.
The observed lack of differences between the three engines suggests that there is at present no indication that LLM-based systems are better at translating metaphors than NMT, at least for literary texts. All three systems struggled with idiomatic expressions, phrasal verbs, and collocations, even when they were highly conventional, frequent, and occurring in their standard form. 

The results from the interviews align with previous studies on literary MT and PE, confirming that post-editing literary MT output is generally perceived as effortful, and as less efficient than translating from scratch. The participants reported frustration, reduced creativity, and a feeling of doubting their own language skills, supporting previous claims in the literature. As far as their responses to metaphor were concerned, the data suggest that when PEs were more aware of the metaphors during the task, they were also more likely to change them and mention them during the interview (i.e. they also remembered them). References to metaphors co-occurred with references to overall poor quality as well as to the output being too literal. Interestingly, a number of PEs indicated that they felt their mind was being “tricked" by the output into accepting translations that were not in fact idiomatic in Dutch, but once they “got stuck in the MT's track", they could no longer think of the right way to say it.

The current paper is of course limited in generalisability, analysing data from only six post-editors for three short excerpts, for one language pair, and comparing only three commercial, black-box systems. Moreover, further research is needed to determine how many of the changes that were made by the PEs were optional rather than required, and how often they retained metaphor translations that could be classified as errors. This will provide more insights into when and how often PEs are “lured into" accepting incorrect MT output. As to why they retain MT metaphors, more research still needs to be done on whether incorrect metaphor translations were in fact not noticed by the post-editors or whether they left the output unchanged because they could not think of an alternative, because they did not care enough to change it given the poor overall quality and poor remuneration, or because they suspected the reader might actually like it and find it creative.   

\section{Acknowledgments}

This publication is part of the project Metaphor in Machine Translation: Reactions, Responses, Repercussions with file number VI.Vidi.231C.014 of the research program Vidi SGW which is (partly) financed by the Dutch Research Council (NWO) under the grant https://doi.org/10.61686/ASYVT63546. 

The data for this publication were collected with financial support from the European Association for Machine Translation (EAMT) through its 2024 sponsorship of activities programme, proposal entitled "Metaphor in Literary Machine Translation and Post-Editing". We would like to thank the post-editors for their participation.

% \bibliography{\confname}
\bibliographystyle{eamt26}
\bibliography{references}

\clearpage
\appendix
\section{Distribution of coded themes across
participants}
\onecolumn
\begin{longtable}{p{6cm}rrrrrrr}

\toprule
Category & PE1 & PE2 & PE3 & PE4 & PE5 & PE6 & Total \\
\midrule
\endfirsthead

\toprule
Category & PE1 & PE2 & PE3 & PE4 & PE5 & PE6 & Total \\
\midrule
\endhead

\midrule
\multicolumn{8}{r}{\textit{Continued on next page}} \\
\endfoot

\bottomrule
\endlastfoot

Efficiency / Time-saving & 8 & 2 & 4 & 6 & 9 & 3 & 32 \\
\quad Faster & 0 & 0 & 1 & 0 & 0 & 0 & 1 \\
\quad More work & 4 & 2 & 3 & 4 & 3 & 1 & 17 \\
\quad Slower & 4 & 0 & 0 & 2 & 7 & 2 & 15 \\

Experience & 8 & 8 & 5 & 17 & 13 & 2 & 53 \\
\quad Difficult & 1 & 2 & 1 & 3 & 0 & 0 & 7 \\
\quad Doubts about language skills & 3 & 4 & 2 & 6 & 7 & 0 & 22 \\
\quad Frustration & 0 & 0 & 0 & 3 & 1 & 0 & 4 \\
\quad Fun & 1 & 1 & 1 & 0 & 1 & 1 & 5 \\
\quad Lack of context & 2 & 1 & 1 & 1 & 3 & 0 & 8 \\
\quad Limits creativity & 2 & 1 & 1 & 7 & 2 & 1 & 14 \\

Fragments & 1 & 1 & 1 & 1 & 1 & 1 & 6 \\
\quad Fragment 2 problematic & 1 & 1 & 1 & 1 & 1 & 0 & 5 \\
\quad No difference & 0 & 0 & 0 & 0 & 0 & 1 & 1 \\

Metaphors & 17 & 8 & 2 & 8 & 1 & 6 & 42 \\
\quad Noticed issues & 16 & 5 & 1 & 8 & 0 & 2 & 32 \\
\quad Noticed positive use & 1 & 1 & 0 & 0 & 0 & 1 & 3 \\
\quad Aware after probing & 0 & 1 & 0 & 0 & 1 & 2 & 4 \\
\quad Not aware after probing & 0 & 1 & 1 & 0 & 0 & 1 & 3 \\

Quality (negative) & 10 & 17 & 4 & 15 & 12 & 9 & 67 \\
\quad Overall issues & 3 & 3 & 3 & 4 & 5 & 0 & 18 \\
\quad Grammar & 0 & 2 & 0 & 0 & 0 & 0 & 2 \\
\quad Too literal & 3 & 6 & 0 & 6 & 5 & 2 & 22 \\
\quad Unnatural language & 2 & 5 & 2 & 4 & 2 & 4 & 19 \\
\quad Vocabulary & 2 & 1 & 0 & 1 & 0 & 3 & 7 \\

Quality (positive) & 9 & 4 & 0 & 0 & 0 & 4 & 17 \\
\quad MT sufficient for PE & 9 & 4 & 0 & 0 & 0 & 4 & 17 \\

Post-editing process & 3 & 2 & 5 & 11 & 1 & 2 & 24 \\
\quad Time/pay affects quality & 2 & 2 & 4 & 3 & 1 & 2 & 14 \\
\quad Full rewriting & 1 & 0 & 0 & 2 & 0 & 0 & 3 \\
\quad Poor source text quality & 0 & 0 & 1 & 7 & 0 & 0 & 8 \\

Style & 0 & 1 & 2 & 5 & 4 & 7 & 19 \\
\quad Fiction problematic for MT & 0 & 0 & 0 & 3 & 3 & 5 & 11 \\
\quad Fiction suitable for MT & 0 & 0 & 1 & 0 & 0 & 2 & 3 \\
\quad Stylistic issues & 0 & 1 & 1 & 3 & 1 & 0 & 6 \\

Training & 4 & 4 & 3 & 3 & 3 & 3 & 20 \\
\quad MT training & 1 & 1 & 1 & 1 & 0 & 1 & 5 \\
\quad MT experience & 1 & 1 & 1 & 0 & 1 & 0 & 4 \\
\quad No MT training & 0 & 0 & 0 & 0 & 1 & 0 & 1 \\
\quad No MT experience & 0 & 0 & 0 & 1 & 0 & 1 & 2 \\
\quad No PE training & 1 & 0 & 1 & 0 & 1 & 0 & 3 \\
\quad No PE experience & 1 & 0 & 0 & 1 & 0 & 1 & 3 \\
\quad PE training & 0 & 1 & 0 & 1 & 0 & 1 & 3 \\
\quad PE experience & 0 & 1 & 1 & 0 & 1 & 0 & 3 \\

\midrule
Total & 121 & 95 & 55 & 138 & 91 & 75 & 575 \\
\multicolumn{8}{p{12cm}}{\footnotesize \textit{Note.} Counts represent coded segments identified in participant responses}
\label{tab:appendix_codes}
\end{longtable}
\twocolumn

\section{Literal English translation of all “to get under his skin" Dutch renderings} 
\begin{table}[ht]
\centering
\begin{tabular}{|p{4.5cm}|p{2cm}|p{6cm}|}
\hline
\textbf{Sentence} & \textbf{Source} & \textbf{Translation (word-by-word English)} \\ \hline

\multirow{8}{4.5cm}{``It's important you understand the background,'' said Sorge, ``\textbf{not allowing the American to get under his skin.''}}
& ChatGPT & Without the American under his skin let crawl \\ \cline{2-3}

& PE1 & Who the American no chance gave to get hold of him \\ \cline{2-3}

& PE2 & Who the American the chance not gave to him to chase \\ \cline{2-3}

& DeepL & Not allowing that the American under his skin \\ \cline{2-3}

& PE3 & Who the American him not on rod let chase \\ \cline{2-3}

& PE4 & Who himself not let chase by the American \\ \cline{2-3}

& GT & While he the American not under his skin let crawl \\ \cline{2-3}

& PE5 & Insensitive to the nagging of the American \\ \cline{2-3}

& PE6 & While he the American not the chance gave to him on the nerves work \\ \hline

\end{tabular}
\label{tab:literal_translations}
\end{table}

\end{document}